\pdfoutput=1

\documentclass[11pt]{article}

\usepackage[final]{acl}

\usepackage{times}
\usepackage{latexsym}

\usepackage[T1]{fontenc}

\usepackage[utf8]{inputenc}

\usepackage{microtype}

\usepackage{inconsolata}

\usepackage{graphicx}
\usepackage{amsmath, amssymb, bm}
\usepackage{algorithm}
\usepackage{multirow}
\usepackage{graphicx}
\usepackage{booktabs}
\usepackage{algorithm}
\usepackage{algorithmic}
\usepackage{multicol}
\usepackage{hyperref}
\usepackage{amsmath}
\usepackage{caption}
\usepackage{enumitem}
\usepackage{wrapfig}
\usepackage{listings}
\usepackage{pifont}
\usepackage{array}
\usepackage{placeins}
\usepackage{float}
\definecolor{codegreen}{rgb}{0.0, 0.411, 0.243}
\definecolor{codered}{rgb}{0.89, 0.26, 0.20}
\usepackage{hyperref}
\definecolor{dartgreen}{HTML}{00693e}
\newcommand{\Checkmark}{\textcolor[rgb]{0.0, 0.0, 0.0}{\ding{51}}}
\newcommand{\XSolidBrush}{\textcolor{black}{\ding{55}}} 
\definecolor{refcolor}{HTML}{9F363A}
\hypersetup{
    colorlinks=true,
    linkcolor=refcolor,
    citecolor=refcolor,
    filecolor=magenta,      
    urlcolor=refcolor,
    }

\title{Learning Musical Representations for\\ Music Performance Question Answering}
\author{
 \textbf{Xingjian Diao\textsuperscript{1}},
 \textbf{Chunhui Zhang\textsuperscript{1}},
 \textbf{Tingxuan Wu\textsuperscript{2}},
 \textbf{Ming Cheng\textsuperscript{1}},
\\
 \textbf{Zhongyu Ouyang\textsuperscript{1}},
 \textbf{Weiyi Wu\textsuperscript{1}},
 \textbf{Jiang Gui\textsuperscript{1}}
\\
 \textsuperscript{1}Dartmouth College
 \\
 \textsuperscript{2}London School of Economics and Political Science
\\
   \texttt{xingjian.diao.gr@dartmouth.edu}
}

\begin{document}
\maketitle

\begin{abstract}
Music performances are representative scenarios for audio-visual modeling.
Unlike common scenarios with sparse audio, music performances continuously involve dense audio signals throughout.
While existing multimodal learning methods on the audio-video QA demonstrate impressive capabilities on general scenarios, they are incapable of dealing with fundamental problems within the music performances: they underexplore the interaction between the multimodal signals in performance and fail to consider the distinctive characteristics of instruments and music.
Therefore, existing methods tend to answer questions regarding musical performances inaccurately. 
To bridge the above research gaps, 
\textit{(i)} given the intricate multimodal interconnectivity inherent to music data, our primary backbone is designed to incorporate multimodal interactions within the context of music;
\textit{(ii)} to enable the model to learn music characteristics, we annotate and release \textit{rhythmic} and \textit{music sources} in the current music datasets;
\textit{(iii)} for time-aware audio-visual modeling, we align the model's music predictions with the temporal dimension. Our experiments show state-of-the-art effects on the Music AVQA datasets. Our code is available at \url{https://github.com/xid32/Amuse}.

\end{abstract}

\section{Introduction}
Imagine standing in a crowd at a live music concert, where every chord strummed, every beat of the drum, and every visual cue from the performers are in perfect harmony, creating an immersive spectacle of sound and sight. Unlike common scenarios with sparse audio signals, music performances offer a tight but fluid fusion of audio and visual elements, ideal for exploring effective audio-visual scene understanding and reasoning \cite{li2022learning, liu2024tackling}, due to the dense and continuous audio signals throughout. This makes modeling music performances one of the most representative tasks in audio-visual learning.

Recently, \citet{li2022learning} introduce Music AVQA, an Audio-Visual Question Answering (AVQA) dataset for music performances. \citet{liu2024tackling} later extend this dataset, refining a more challenging and balanced version named Music AVQA-v2. Music AVQA datasets require models to leverage both audio and visual representations to answer questions that closely correspond to the presented audio-visual content.

\begin{figure}[tb]
\centering
\resizebox{0.48\textwidth}{!}{
\includegraphics{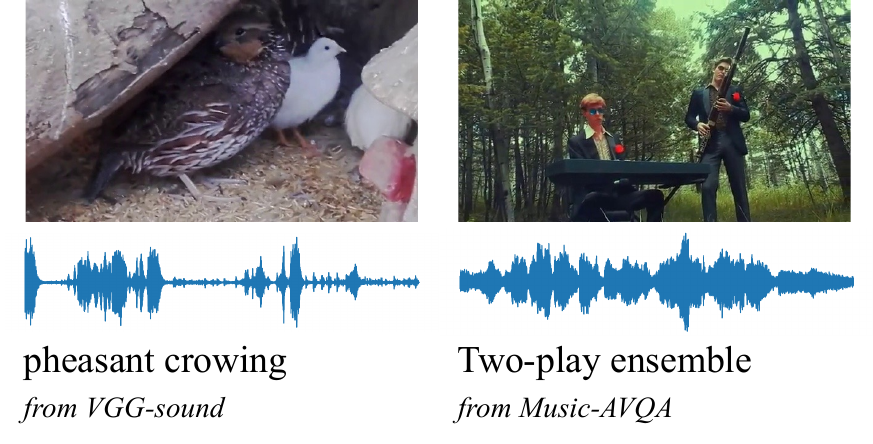}
}
\caption{Sparse audio in general videos vs. continuous audio in music performance videos: The left natural video has sparse audio signals (occasional chirping) \cite{9053174}.
The right is a sample from the Music AVQA, showing the dense and continuous audio signals of music performances \cite{li2022learning}.}
\label{fig:teaser}
\vspace{-0.5cm}
\end{figure}

Existing deep learning methods for AVQA have made notable progress: earlier research by \citet{ngiam2011multimodal} showcase cross-modality feature learning to boost audio-visual speech recognition, while \citet{srivastava2012multimodal} use a multimodal Deep Boltzmann Machine for multimodal learning. More recent strategies in multimodal fusion \cite{yun2021pano, yang2022avqa} and creating positive-negative pairs \cite{li2022learning} have been introduced to improve audio-visual correlation learning. As a later advancement, LAVisH \cite{lin2023vision} is introduced as an adapter to extend the pretrained ViT \cite{dosovitskiy2020image} to audio-visual data for cross-modal learning. Additionally, \citet{duan2024cross} present DG-SCT, which uses audio and visual modalities as prompts in frozen pretrained encoders.

However, prior AVQA methods, designed for general purposes, face significant challenges when applied to specific music performances, which involve continuously dense audio and interconnected multimodal music signals (as shown in Fig.\ref{fig:teaser}). These challenges result in sub-optimal accuracy:
\textit{{(i)}} they often overlook the musical characteristics (rhythm and music source) of instruments and players, as their designs are not originally specified for music performances;
\textit{{(ii)}} they lack sufficient explicit supervision for modeling music rhythm and music sources over time, which is essential for answering temporal-related questions. Most methods rely on attention mechanisms but fail to directly align music rhythm and music sources to the temporal dimension for audio-visual QA.
\textit{{(iii)}} more essentially, these methods often use either unimodal encoders or parallel non-interactive encoders. This results in influence by irrelevant noise when processing other modalities and specific questions, or a lack of early-stage fusion on intricate interconnected multimodal representations for music AVQA data.
To overcome these challenges, we propose \textbf{\texttt{Amuse}} framework, differing from prior works as shown in Tab.\ref{tab:baseline-compare}, specially designed for \underline{\textbf{A}}udio-{{V}}isual {{Q}}uestion {{A}}nswering in \underline{\textbf{mus}}ical performanc\underline{\textbf{e}}:
\begin{itemize}[leftmargin=*]
\item To fully exploit the intricate interconnectivity of multimodal music data (continuous audio, video, and text), we design the multimodal interactive encoder to facilitate interactions between tokens from different modalities throughout the forward.

\item For musical characteristics explicitly learnable to models, we annotate and release music rhythm and sources in the Music AVQA dataset.

\item To ensure the framework's time-awareness in processing audio-visual modalities for temporal-musical questions, we designed rhythm and music source encoders that explicitly align musical signals to the temporal dimension.
\end{itemize}

\begin{table}[t]
\centering
\resizebox{0.495\textwidth}{!}{
\begin{tabular}{lccccc}
\toprule
\multirow{2}{*}{{AVQA}} & \multicolumn{5}{c}{{Musical Designs}} \\
                         & {avq} & {rhy.} & {src.} & {temp.} & {g.l.f.} \\ \midrule
AVST$^{1}$ & \XSolidBrush & \XSolidBrush & \XSolidBrush & \XSolidBrush & \XSolidBrush \\
PSTP-Net$^{2}$ & \XSolidBrush & \XSolidBrush & \XSolidBrush & \XSolidBrush  & \XSolidBrush \\
LAVisH$^{3}$  & \XSolidBrush   & \XSolidBrush & \XSolidBrush & \XSolidBrush & \XSolidBrush \\
LSTTA$^{4}$  & \Checkmark   & \XSolidBrush   & \XSolidBrush & \XSolidBrush & \XSolidBrush \\
DG-SCT$^{5}$  & \XSolidBrush  & \XSolidBrush & \XSolidBrush & \XSolidBrush & \XSolidBrush \\
LAST-Att$^{6}$  & \XSolidBrush   & \XSolidBrush & \XSolidBrush & \XSolidBrush & \XSolidBrush \\ 
\textbf{\texttt{Amuse}}                     & \Checkmark   & \Checkmark   & \Checkmark   & \Checkmark   & \Checkmark   \\
\bottomrule
\end{tabular}
}
\label{dataset}
\begin{minipage}{0.5\textwidth}
\caption*{
\footnotesize
$^{1}$ \citet{li2022learning}
$^{2}$ \citet{li2023progressive}
$^{3}$ \citet{lin2023vision}
\\
$^{4}$ \citet{liu2023parameter}
$^{5}$ \citet{duan2024cross}
$^{6}$ \citet{liu2024tackling}
}
\end{minipage}
\vspace{-0.6cm}
\caption{Comparison of \texttt{Amuse} with previous SoTA AVQA studies on characteristics of music performance:
\textit{avq} means audio-visual-question interactions from early fusion (for dense audio); 
\textit{rhy.} indicates rhythm integration; 
\textit{src.} refer to music source integration; 
\textit{Temp.} denotes temporal alignment on music; 
and \textit{g.l.f} means global features and local musical regions of interest.}
\vspace{-0.4cm}
\label{tab:baseline-compare}
\end{table}

\section{Related Work}

\paragraph{Audio-Visual Scene Understanding}
Audio-visual scene understanding tasks leverage audio and video features to interpret the surrounding environment, gaining significant popularity in the computer vision community \cite{antol2015vqa, zhao2018sound, zhao2019sound, tian2020unified}. Recently, to enhance audio-visual correlation learning, multi-modal fusion \cite{yun2021pano, yang2022avqa, diao2023av} and positive-negative pair construction \cite{li2022learning} methods are proposed. State-of-the-art (SoTA) methods include LAVisH \cite{lin2023vision}, an adapter that generalizes pretrained ViT \cite{dosovitskiy2020image} to audio-visual data for cross-modal learning, and DG-SCT \cite{duan2024cross}, which leverages audio and visual modalities as prompts in pretrained frozen encoders.
Despite advances, fundamental problems persist, including temporal inconsistency between audio-visual modalities and inaccurate question answering. To address these challenges, we propose \texttt{Amuse}, which identifies musical specialties like rhythm and music sources from both vision and audio, and further aligns them with the temporal dimension.

\paragraph{Question Answering by AI Models}
Visual Question Answering (VQA) \cite{antol2015vqa, lei2018tvqa, yu2019activitynet, garcia2020knowit, ravi2023vlc, yu2024self} and Audio Question Answering (AQA) \cite{fayek2020temporal, lipping2022clotho, sudarsanam2023attention, li2023multi} have been widely studied for scene understanding. These methods leverage features from either the visual or audio modality to learn relations with linguistic features, often omitting audio-visual cross-modality. Recently, Audio-Visual Question Answering (AVQA) has been introduced \cite{yun2021pano, li2022learning}, which requires models to use both audio and visual features to answer questions. Initially, \citet{yun2021pano} proposes Pano-AVQA, an audio-visual question-answering dataset for panoramic videos, while \citet{li2022learning} creates the Music AVQA benchmark and introduces a spatio-temporal grounded audio-visual network. Afterward, LAVisH \cite{lin2023vision} and DG-SCT \cite{duan2024cross} are proposed for scene understanding through audio-visual cross-modality. As the latest work, \citet{liu2024tackling} extends the Music AVQA dataset and introduces a new baseline model. Recently, with the development of large language models, explorations of involving LLMs for questioning/answering have attracted widespread attention across various applications \cite{zhuang2023toolqa, wang2024healthq, saito2024unsupervised}. Specifically, for applying audio modalities to language, AudioGPT \cite{huang2024audiogpt} and HuggingGPT \cite{shen2024hugginggpt} use well-trained foundational audio models as tools while utilizing LLMs as flexible interfaces. Besides, explorations about transferring language modeling techniques to music generation have been introduced \cite{copet2024simple, dai2022missing, agostinelli2023musiclm, lu2023musecoco}. These studies prove that symbolic music can be treated similarly to natural language, and further utilize the pretrained knowledge from LLMs to enhance music generation. Unlike the aforementioned directions, our study focuses on semantic understanding and classification in the Music-AVQA task, where we annotate \textit{rhythm} and \textit{source} information in existing music datasets to make musical characteristics explicitly learnable.

\section{\texttt{Amuse} Framework}
\begin{figure*}
\begin{center}
\resizebox{1\textwidth}{!}{
    \includegraphics{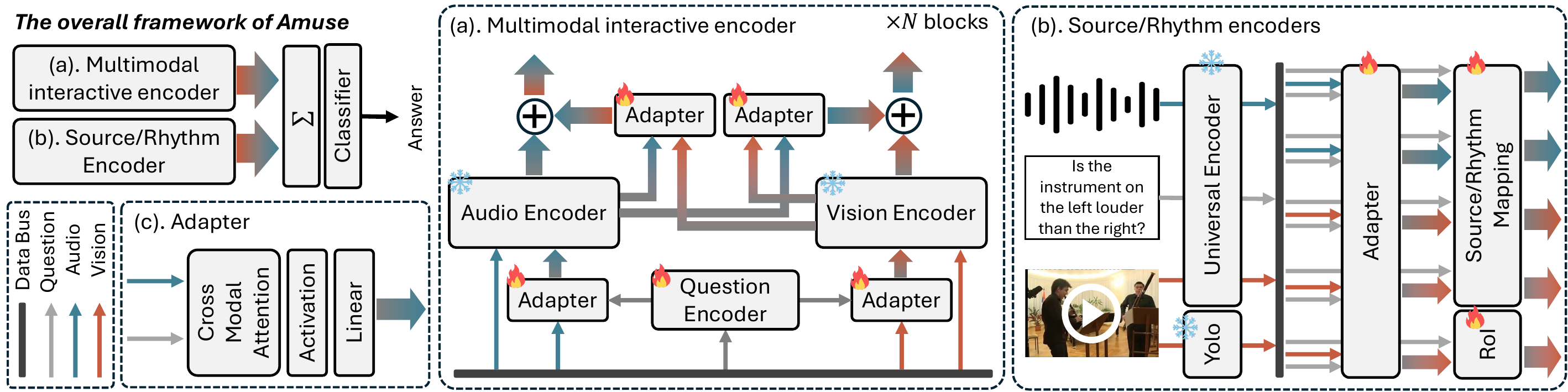}
}
\end{center}
\caption{\texttt{Amuse} framework integrates (a) and (b) combining multimodal interactive and musical-specialized representations for answering: (a) multimodal interactive encoder's audio, vision, and question modules are interconnected via adapters that perform cross-modal attention. (b) Source/Rhythm encoders extract and encode musical-specific characteristics such as rhythm and sound sources. Universal Encoder incorporates pretrained vision and audio source/rhythm encoders as depicted in Fig.\ref{fig:predictor}. A musical RoI extractor (light weight Yolo) detects music-related elements like instruments and performers. They are aligned with the temporal dimension. 
}
\vspace{-0.3cm}
\label{fig:amuse}
\end{figure*}
\subsection{Interacting Multimodal Representations}
\texttt{Amuse} deviates from the conventional use of unimodal or parallel non-interactive encoders, which often result in irrelevant noise and fail to capture complex multimodal music representations, such as those with non-fusion or late fusion making insufficient interactions. Instead, we use three interactive transformers that dynamically interact with each other \textit{throughout} the forward processing of inputs from three modalities in music-AVQA: 

Swin-V2 transformer \cite{liu2022swin} processes music concert video inputs, the Hierarchical Token-Semantic Audio Transformer (HTS-Audio Transformer) \cite{chen2022hts} processes music audio inputs, and the language transformer \cite{NIPS2017_3f5ee243} processes music-related questions.
Then, to facilitate interaction between multimodal representations from an early stage, cross-modal adapters are used between the three transformers. These adapters consist of cross-modal attention operating on tokens from two different modalities, followed by a linear layer with activation to project the output tokens into the next blocks of each transformer.

In detail, let $\mathbf{X}_v \in \mathbb{R}^{T_v \times d}$ represent the visual tokens from the Swin-V2 transformer, $\mathbf{X}_a \in \mathbb{R}^{T_a \times d}$ represent the audio tokens from the HTS-Audio Transformer, and $\mathbf{X}_l \in \mathbb{R}^{T_l \times d}$ represent the language tokens from the language transformer, where $T_v$, $T_a$, and $T_l$ denote the token lengths and $d$ denotes the dimensionality of the token embeddings. The cross-modal attention mechanism enables interaction between tokens from different modalities. For simplicity, consider cross-modal attention between visual and audio tokens. The query, key, and value matrices are derived as $\mathbf{Q}_v = \mathbf{X}_v \mathbf{W}_Q$, $\mathbf{K}_a = \mathbf{X}_a \mathbf{W}_K$, and $\mathbf{V}_a = \mathbf{X}_a \mathbf{W}_V$,
where $\mathbf{W}_Q$, $\mathbf{W}_K$, and $\mathbf{W}_V$ are learnable weight matrices. The attention scores and the attended output $\mathbf{A}_{va}$ are:
\begin{equation}
\mathbf{A}_{va} = \text{softmax}\left(\frac{\mathbf{Q}_v \mathbf{K}_a^T}{\sqrt{d}}\right) \mathbf{V}_a.
\end{equation}
The cross-modal adapter takes the attended output $\mathbf{A}_{va}$ and projects it into the next block of the visual transformer:
\begin{equation}
\mathbf{Z}_v = \sigma(\mathbf{A}_{va} \mathbf{W}_1 + \mathbf{b}_1) \mathbf{W}_2 + \mathbf{b}_2,
\end{equation}
where $\mathbf{W}_1$ and $\mathbf{W}_2$ are learnable weight matrices, $\mathbf{b}_1$ and $\mathbf{b}_2$ are biases, and $\sigma$ is an activation function (e.g., ReLU).

The same process can be applied to enable interactions between any pair of modalities (e.g., visual-audio, visual-language, audio-language). Therefore, the final fused representation for a modality is given by combining the attended outputs from different cross-modal adapters:
\begin{equation}
\begin{aligned}
\mathbf{X}_v^{fused} &= \text{Fuse}(\mathbf{X}_v, \mathbf{A}_{va}, \mathbf{A}_{vl}), \\
\mathbf{X}_a^{fused} &= \text{Fuse}(\mathbf{X}_a, \mathbf{A}_{av}, \mathbf{A}_{al}), \\
\mathbf{X}_l^{fused} &= \text{Fuse}(\mathbf{X}_l, \mathbf{A}_{lv}, \mathbf{A}_{la}),
\end{aligned}
\end{equation}
where $\text{Fuse}$ denotes the fusion operation, which is the cross-modal attention. By implementing these cross-modal interactions, \texttt{Amuse} effectively captures and integrates complex multimodal representations, facilitating more accurate and nuanced understandings of music-AVQA data.

Thus, each transformer processes its distinctive modal representation with a modal-specific design, while the cross-modal adapters enable the exchange of fused cross-modal tokens, facilitating multimodal representation interactions for the music-AVQA data.

\subsection{Aligning Rhythm and Music Source to the Temporal Dimension}
Music performance data comprise three characteristics distinct from other general audio-vision-question data: music rhythm, music source, and music objects, which are all essential to respond to the questions and convolutedly correlate with audio and vision. 
We first introduce how we annotate the musical rhythm and source distributed along the temporal dimension. 
We then describe the process of learning the temporal representations of annotated musical rhythm and source, then strengthen the previously discussed multimodal interactive pipeline. Finally, we incorporate an object detector to extract local features from musical regions of interest (Musical RoIs). This detector identifies key elements such as instruments, players, and conductors within the video, which are helpful to Music AVQA answer predictions.

\paragraph{Temporal Annotation of Rhythm and Source}
\underline{\textit{(i)}} We designed a rhythm annotation method to identify and annotate temporal segments in an audio signal where significant changes in rhythm occur. This method segments the audio $\mathbf{A}$ into smaller parts (6-second intervals in a 60-second audio/video clip). The audio of length $T$ is divided into $n$ non-overlapping segments, each of length $t$.
For segment $i$, the audio snippet $\mathbf{a}_i$ is defined as:
\begin{equation}
\mathbf{a}_i = \mathbf{A}[i \cdot t : (i+1) \cdot t].
\end{equation}
We use a beat tracking tool $\mathbf{BT}(\cdot)$ to measure the beats per minute (BPM) for each segment as
$\mathbf{bpm}_i = \mathbf{BT}(\mathbf{a}_i)$.
The method calculates the average BPM $\mu_{bpm}$ of the entire audio and sets a threshold $\vartheta$ at 25\% of this average to determine what constitutes a significant change in rhythm:
\begin{equation}
\mu_{bpm} = \mathbf{BT}(\mathbf{A}), \quad \vartheta = 0.25 \times \mu_{bpm}.
\end{equation}
By comparing the BPMs of consecutive segments, the method labels segments where the BPM difference exceeds the threshold, indicating a notable rhythm variation:

\vspace{-0.5cm}
\begin{equation}
\mathbf{Label}_i = 
\begin{cases} 
1 & \text{if } |\mathbf{bpm}_i - \mathbf{bpm}_{i+1}| > \vartheta, \\
0 & \text{otherwise}.
\end{cases}
\end{equation}
The result is a sequence of binary labels that mark where significant rhythm changes happen throughout the audio.

\underline{\textit{(ii)}} Our source segment annotation method focuses on detecting and annotating the presence of specific musical instruments in a temporal audio recording. This method divides the audio into segments (6-second intervals in a 60-second audio/video clip) and uses a Universal Source Separation model \cite{kong2023universal} to identify different music sources within each segment. For segment $i$, the audio snippet $\mathbf{a}_i = \mathbf{A}[i \cdot t : (i+1) \cdot t]$ is processed to obtain the sources as $\mathbf{USS}(\mathbf{a}_i)$.
It counts how often the specified instruments appear in these segments. Counts track the occurrences. By tracking the occurrences of these instruments, the method creates a timeline of annotations indicating when and how frequently each instrument is present throughout the audio.

\begin{figure}
\begin{center}
\resizebox{0.5\textwidth}{!}{
    \includegraphics{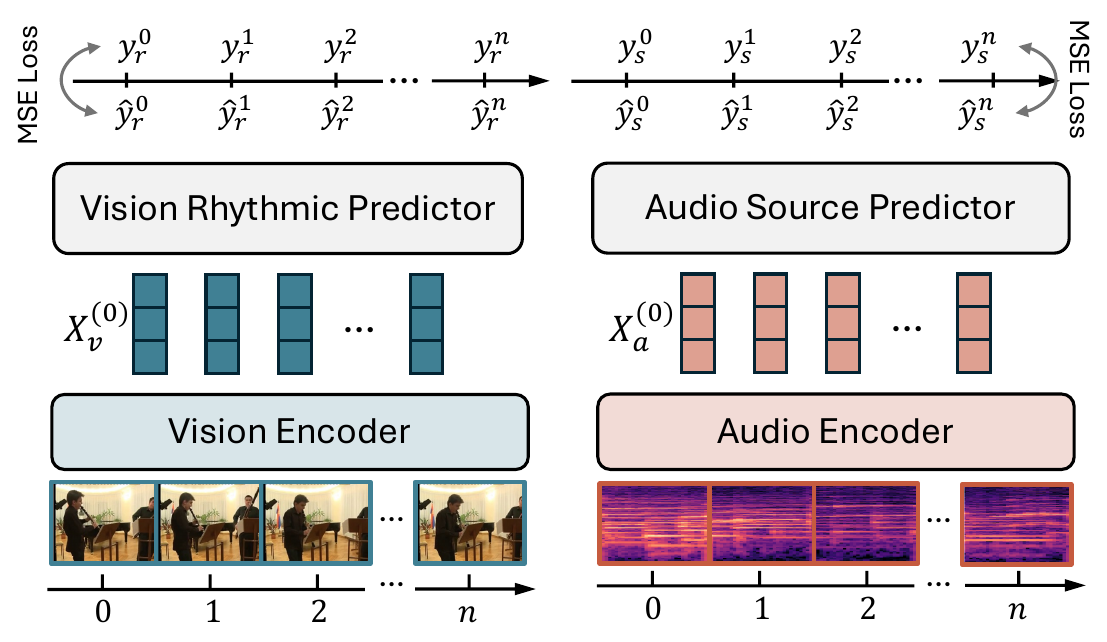}
}
\end{center}
\vspace{-0.5cm}
\caption{Visual frames and audio spectrograms are processed by rhythm and source encoders—one rhythm encoder and one source encoder for each modality (vision and audio)—to align their characteristics along the temporal dimension.}

\vspace{-0.5cm}
\label{fig:predictor}
\end{figure}

\paragraph{Learning Rhythm \& Source over Time}
As illustrated in Fig.\ref{fig:predictor}, our modeling approach processes the annotated rhythm and source data to learn their representations and aid the multimodal interactive encoder.
For processing inputs, musical video are processed through the Swin-V2 Transformer Encoder to generate visual tokens, while musical audio inputs are processed through the HTS-Audio Transformer to produce audio tokens:
\begin{equation}
\begin{aligned}
\mathbf{X}_v^{(l_{\text{layer}})} &= \text{Swin-V2}(\mathbf{X}_v^{(l_{\text{layer}}-1)}), \\
\mathbf{X}_a^{(l_{\text{layer}})} &= \text{HTS-AT}(\mathbf{X}_a^{(l_{\text{layer}}-1)}).
\end{aligned}
\end{equation}
where $l_{\text{layer}}$ denotes the layer index of the encoder.
Then, for average pooling, we apply average pooling to simplify the representations:
\begin{equation}
\begin{aligned}
\mathbf{X}_v^{(avg)} &= \text{avgpool}(\mathbf{X}_v^{(l_{\text{layer}})}), \\
\mathbf{X}_a^{(avg)} &= \text{avgpool}(\mathbf{X}_a^{(l_{\text{layer}})}).
\end{aligned}
\end{equation}
Finally, we pass the pooled representations through rhythm and source predictors, each consisting of three-layer linear networks as predictors:
\begin{equation}
\begin{aligned}
\mathbf{\hat{Y}}_{r} &= \text{Predictor}_{r}(\mathbf{X}_v^{(avg)}, \mathbf{X}_a^{(avg)}),  \\
\mathbf{\hat{Y}}_{s} &= \text{Predictor}_{s}(\mathbf{X}_v^{(avg)}, \mathbf{X}_a^{(avg)}).
\end{aligned}
\end{equation}
When we train the Rhythm and Source encoders, we use Mean Squared Error (MSE) loss with the annotated rhythm and source labels:
\begin{equation}
\mathcal{L}_{MSE} = \frac{1}{N} \sum_{i=1}^{N} \left[ ({\hat{y}}^{i}_{r} - {{y}}^{i}_{r})^2 + ({\hat{y}}^{i}_{s} - {{y}}^{i}_{s})^2 \right].
\end{equation}

After pretraining, these encoders are frozen and used as pretrained models to fulfill the functionality of the Universal Encoder as depicted in Fig.\ref{fig:amuse} (b).

\subsection{Highway for Musical Regions of Interest}
In addition to the primary multimodal interactive encoder and rhythm \& source encoders, we also incorporate an object detector, such as the lightweight YoloV8~\cite{Ultralytics2023yolov8}, to identify musical regions of interest in the video input. This detector captures explicit local features of music-related elements, including instruments, players, and conductors. These local features work as highway/shortcut information that, together with other encoders (for global representations of musical concerts), contribute to the final representations fed into the final layer of \texttt{Amuse} for answer prediction.

Overall, \texttt{Amuse} emphasizes temporal-aware modeling of rhythmic and music sources aligning with the time dimension. The pretrained rhythm and source encoders, which provide musical characteristic features, operate in parallel with a multimodal interactive encoder and highway for musical RoIs, which offers global multimodal features and local musical element features. This collaborative setup ensures comprehensive understanding and accurate answers in music-AVQA tasks.

\section{Experiments}
\texttt{Amuse} addresses unique music characteristics not considered by prior state-of-the-art AVQA methods, as Tab.\ref{tab:baseline-compare} summarized. Thus, first, we compare \texttt{Amuse} with other SoTA AVQA methods to evaluate its performance. Second, we iteratively verify the effectiveness of our multimodal interactive encoder's early fusion on overall performance. We then assess the impact of incorporating music rhythm and source, their alignment with the temporal dimension, and the contribution of musical RoIs to the framework's ability to accurately capture music elements for final predictions.

\subsection{Setup}
\paragraph{Music AVQA dataset: v1 and v2}
\underline{\textit{(i)}} Music AVQA v1 \cite{li2022learning} is designed for multimodal scenarios. It comprises 9,288 videos featuring 22 different musical instruments, totaling 150 hours and containing 45,867 question-answer (QA) pairs. On average, each video contains approximately five QA pairs. 
It includes 33 question templates across 4 categories: String, Wind, Percussion, and Keyboard.
\underline{\textit{(ii)}} Music AVQA v2 \cite{liu2024tackling} further addresses data bias issues in the original Music AVQA dataset. It is curated by manually collecting 1,230 instrument performance videos and creating 8,100 new QA pairs. 

\paragraph{Backbones}
Across all encoders, those taking visual features as input adopt the Swin-v2-large architecture \cite{liu2022swin}, while those taking audio features as input adopt the HTS-AT architecture \cite{chen2022hts}. 
Specifically, the visual encoder consists of 18 layers, while the audio encoder comprises six layers. 
Therefore, a cross-modal adapter is introduced every three layers to handle inputs from the two encoders. 
In addition, a single convolution layer is utilized for dimension transformation in cases where the dimensions do not match upon inputting into the cross-modal adapters. 

\begin{table*}[htbp]
\centering
\resizebox{2.0\columnwidth}{!}{%
\begin{tabular}{@{}l|ccc|ccc|cccccc|c@{}}
\toprule
\multirow{2}{*}{{Methods}}            & \multicolumn{3}{c|}{{Audio-related QA}}              & \multicolumn{3}{c|}{{Visual-related QA}}             & \multicolumn{6}{c|}{{Audio\&Visual-related QA}}                                                          & \multirow{2}{*}{{Avg}} \\
                                             & {Count} & {Comp}  & {Avg}   & {Count} & {Local} & {Avg}   & {Exist} & {Count} & {Local} & {Comp}  & {Temp}  & {Avg}   &                               \\ \midrule
AVST \cite{li2022learning}  & 77.78          & 67.17          & 73.87          & 73.52          & 75.27          & 74.40          & 82.49          & 69.88          & 64.24          & 64.67          & 65.82          & 69.53          & 71.59                         \\
PSTP-Net  \cite{li2023progressive}                                   & 73.97          & 65.59          & 70.90          & 77.15          & 77.36          & 77.26          & 76.18          & 73.23          & 71.80          & 71.19          & 69.00          & 72.57          & 73.52                         \\
LAVisH   \cite{lin2023vision}      & 75.59          & \textbf{84.13} & 76.86          & 77.45          & 72.91          & 76.29          & 71.91          & 77.52          & \underline{75.81}          & 76.75          & 77.62          & 76.31          & 76.10                         \\
LSTTA    \cite{liu2023parameter}                                    & 81.75          & 82.04         & \underline{81.90}          & 81.82          & 82.23          & 82.03          & \underline{83.46}          & \underline{79.11}          & \textbf{78.23} & \underline{78.02}          & \underline{79.32}          &  \underline{79.63}          & \underline{81.19}                         \\
DG-SCT \cite{duan2024cross} & 83.27          & 64.56          & 76.34          & 81.57          & 82.57          & \underline{82.08}          & 81.61          & 72.84          & 65.91          & 64.22          & 67.48          & 70.56          & 74.62                         \\ 
LAST-Att      \cite{liu2024tackling}
& \textbf{85.71}         & 63.10          & -          & \underline{83.86}         & \underline{83.09}          & -          & 76.47          & 76.20          & 68.91          & 65.60          & 66.75          & -          & 75.45     
\\ \midrule
\textbf{\texttt{Amuse} (ours)}                         & \underline{84.61} & \underline{82.45}          & \textbf{83.58} & \textbf{87.14} & \textbf{84.39} & \textbf{85.84} & \textbf{86.95} & \textbf{85.49} & 73.01          & \textbf{82.98} & \textbf{83.06} & \textbf{82.43} & \textbf{83.52}                \\ \bottomrule
\end{tabular}
}
\vspace{-0.1cm}
\caption{Comparison with state-of-the-art methods on the Music AVQA \cite{li2022learning} test set. We report the accuracy for Audio (Counting, Comparative), Visual (Counting, Location), and Audio-Visual (Existential, Counting, Location, Comparative, Temporal) question types, along with the average accuracy for Audio, Visual, Audio-Visual, and overall. \textbf{Bold} results indicate the best performance among {all} methods, while \underline{underlined} results indicate the second-best performance among {all} methods. \texttt{Amuse} results in all tables are averaged over three runs.}
\label{tab:tab1}
\end{table*}

\begin{table*}[htbp]
\centering
\resizebox{2\columnwidth}{!}{%
\begin{tabular}{@{}l|ccc|ccc|cccccc|c@{}}
\toprule
\multirow{2}{*}{{Methods}} & \multicolumn{3}{c|}{{Audio-related QA}}              & \multicolumn{3}{c|}{{Visual-related QA}}             & \multicolumn{6}{c|}{{Audio\&Visual-related QA}}                                                          & \multirow{2}{*}{{Avg}} \\
                                  & {Count} & {Comp}  & {Avg}   & {Count} & {Local} & {Avg}   & {Exist} & {Count} & {Local} & {Comp}  & {Temp}  & {Avg}   &                               \\ \midrule
AVST      \cite{li2022learning}                         & 81.74          & 62.11          & 72.46          & 79.08          & 77.64          & 78.40          & 72.12          & 69.03          & 65.05          & 63.98          & 60.57          & 66.26          & 71.08                         \\
LAVisH       \cite{lin2023vision}                     & 84.36          & 58.57          & 72.17          & 83.25          & 81.46          & 82.40          & 73.26          & 73.45          & 65.64          & 64.26          & 60.82          & 67.75          & 72.34                         \\
DG-SCT     \cite{duan2024cross}                      & 83.66          & 62.47          & \underline{73.64}          & 82.05          & 82.97          & \underline{82.48}          & \underline{83.43}          & 72.70          & 64.65          & 64.78          & \underline{67.34}          & \underline{70.38}          & 74.08                         \\
LAST-Att      \cite{liu2024tackling}                    & \textbf{86.03}          & \underline{62.52}          & -              & \underline{84.12}          & \underline{84.01}          & -              & 76.21          & \underline{75.23}          & \underline{68.91}          & \underline{65.60}          & 60.60          & -              & \underline{75.44}                         \\ \midrule
\textbf{\texttt{Amuse} (ours)}              & \underline{84.76} & \textbf{83.88} & \textbf{84.34} & \textbf{88.15} & \textbf{85.16} & \textbf{86.74} & \textbf{88.30} & \textbf{87.47} & \textbf{78.77} & \textbf{84.41} & \textbf{85.38} & \textbf{85.51} & \textbf{85.16}                \\ \bottomrule
\end{tabular}
}
\vspace{-0.1cm}
\caption{Comparison with state-of-the-art methods on the Music AVQA-v2 \cite{liu2024tackling} test set.}
\vspace{-0.4cm}
\label{tab:tab2}
\end{table*}

\paragraph{Training configurations}
During training, we employ the Adam optimizer \cite{kingma2015adam}. The training process for \texttt{Amuse} is divided into two stages with distinct configurations:
For the pretraining on sub-parts of \texttt{Amuse}, specifically the rhythm and music source predictors, we use a learning rate of 0.001. This stage exclusively utilizes the rhythm and music source annotations to pretrain these predictors.
During finetuning the full pipeline of \texttt{Amuse}, we begin with an initial learning rate of 0.0001; the learning rate is reduced by 50\% every 5 epochs to ensure gradual convergence. 

\subsection{Overall Comparisons}
\label{subsec:exp-3}
Tab.\ref{tab:tab1} and \ref{tab:tab2} present a comprehensive comparison of various methods on the Music AVQA and Music AVQA-v2 test set. The evaluation metric used is accuracy. Questions are categorized into three groups: audio-visual, audio, and visual, based on their primary focus. \texttt{Amuse} demonstrates impressive performance across all categories on both v1 and v2 datasets. Note that all experiments' results reported for \texttt{Amuse} are averaged over three runs.

\textbf{Audio-Visual Question Answering.} For the question related to audio-visual joint modalities, \texttt{Amuse} consistently exhibits significantly remarkable performance, with the highest accuracy among all baselines. Specifically, \texttt{Amuse} achieves an accuracy of 85.49\% for audio-visual counting questions, which leads to the current best baseline by at least 6.38\% \cite{liu2023parameter}. Similar results can be observed for temporal questions (at least +3.74\% against \cite{liu2023parameter}) and the overall performance (at least +2.33\% against \cite{liu2023parameter}). 

\begin{figure}[htp]
\begin{center}
\includegraphics[width=1.0\linewidth]{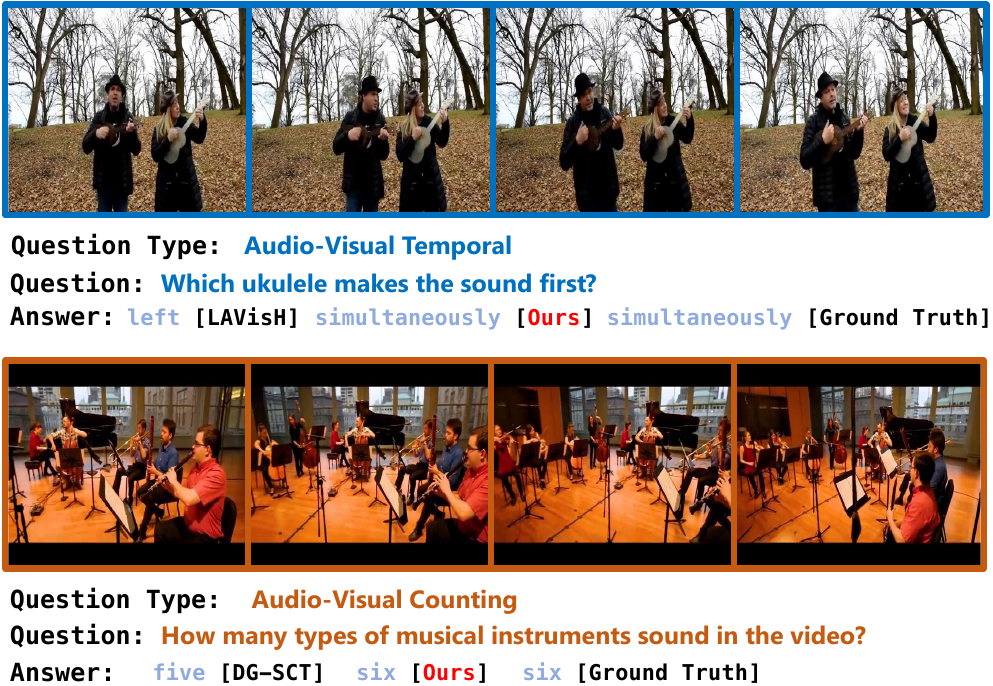}
\end{center}
\vspace{-0.3cm}
\caption{Demonstration of audio-visual \textit{temporal} and \textit{counting} QA. We show examples that our model correctly handles audio-visual temporal and counting questions, while SoTA models LAVisH and DG-SCT fail.}
\label{fig:twoquestion}
\vspace{-0.2cm}
\end{figure}

Additionally, Fig.\ref{fig:twoquestion} illustrates two instances where our model effectively tackles audio-visual temporal and counting questions, contrasting with the shortcomings of state-of-the-art models LAVisH and DG-SCT. For the complex temporal question (same musical instrument in the video shown at the top), our model precisely identifies the temporal sequence of the instrument sounds. Furthermore, our model accurately provides the correct answer for the counting question amidst noise (six instruments playing in the video shown at the bottom).
\begin{table*}[t]
\centering
\resizebox{2\columnwidth}{!}{%
\begin{tabular}{@{}l|ccc|ccc|cccccc|c@{}}
\toprule
\multirow{2}{*}{{Methods}} & \multicolumn{3}{c|}{{Audio-related QA}}              & \multicolumn{3}{c|}{{Visual-related QA}}             & \multicolumn{6}{c|}{{Audio\&Visual-related QA}}                                                          & \multirow{2}{*}{{Avg}} \\
                                  & {Count} & {Comp}  & {Avg}   & {Count} & {Local} & {Avg}   & {Exist} & {Count} & {Local} & {Comp}  & {Temp}  & {Avg}   &                               \\ \midrule
w/o M.i.e.                       & 68.58          & 56.06          & 62.66          & 53.48          & 53.02          & 53.26          & 47.07          & 62.53          & 56.11          & 50.05          & 43.92          & 51.54          & 54.73                         \\
w/o Rhy.                      & 83.21          & 80.17          & 81.77          & 76.95          & \textbf{85.08}          & 80.79          & 86.28          & 84.06          & 72.52          & 83.47          & \textbf{83.59}          & \textbf{84.08}          & 82.72                         \\
w/o Src.                        & 78.67          & 79.08          & 78.86          & 84.64          & 82.69          & 83.72          & 85.98          & 79.94          & 71.58          & 83.19          & 80.67          & 80.26          & 80.73                         \\
w/o RoIs                          & 83.27          & \textbf{82.62}          & 82.96          & 80.26          & 76.43          & 78.45          & 80.95          & 72.09          & 72.63          & \textbf{84.61}          & 83.30          & 79.06          & 79.89                         \\ \midrule
\textbf{\texttt{Amuse} (ours)}                         & \textbf{84.61} & 82.45          & \textbf{83.58} & \textbf{87.14} & 84.39 & \textbf{85.84} & \textbf{86.95} & \textbf{85.49} & \textbf{73.01}          & 82.98 & 83.06 & 82.43 & \textbf{83.52}                \\ \bottomrule
\end{tabular}
}
\caption{Ablation studies on the Music AVQA \cite{li2022learning} test set. w/o M.i.e. means without the multimodal interactive encoder; w/o Rhy. means without the rhythm annotation module; w/o Src. means without the source annotation module; w/o M. RoIs means without the highway for Music Regions of Interest (RoIs).
}
\vspace{-0.4cm}
\label{tab:tab3}
\end{table*}

\textbf{Audio Question Answering.} For the questions specifically related to audio modality, \texttt{Amuse} achieves competitive performance against SoTA models, where our model ranks first/second across all metrics on Music AVQA test set.  Specifically, for counting questions, \texttt{Amuse} achieves an accuracy of 84.61\%, which is competitive against the latest proposed method LAST-Att \cite{liu2024tackling}. Moreover, for audio-comparative questions, our model leads LAST-Att by remarkably 19.35\%. Similar observations can be concluded when compared to other recently proposed models \cite{liu2023parameter, lin2023vision, duan2024cross}. 

In addition, our model achieves an average accuracy of 83.58\% in total, demonstrating its robust performance in various audio-related tasks and its effectiveness in handling audio-specific questions. Similar findings can be observed for the Music AVQA-v2. \texttt{Amuse} achieves an accuracy of 84.76\%, 83.88\%, and 84.34\% for three categories, strongly demonstrating its effectiveness by its best average audio performance.

\textbf{Video Question Answering.} For the questions specifically related to visual modality, \texttt{Amuse} consistently demonstrates superior performance compared to other methods on Music AVQA test set. For visual counting and location questions, \texttt{Amuse} achieves an accuracy of 87.14\% and 84.39\%, outperforming all other methods. Moreover, the average accuracy of \texttt{Amuse} is 85.84\%, indicating its effectiveness in visual-specific tasks. Similar results can be evidently observed for Music AVQA-v2, where \texttt{Amuse} outperforms all related work under all evaluation metrics. 

Overall, as evidenced by Tab.\ref{tab:tab1} and \ref{tab:tab2}, \texttt{Amuse} consistently surpasses other methods across various question types and categories, demonstrating its capability in tackling the challenges presented by the Music AVQA and Music AVQA-v2 datasets. With its strong performance and high accuracy, \texttt{Amuse} stands out as a promising solution for audio-visual question-answering tasks for musical performance, because of the effectiveness of our design: multimodal representations interactions (across audio, visual, and question),  aligning musical rhythm \& music source with temporal dimension, and highway for musical regions of interest. 
\subsection{Impacts of Each Musical Design}
\label{subsec:exp-5}

We conducted ablation studies to remove musical-related modules from the \texttt{Amuse} framework and assessed their impact, as shown in Tab.\ref{tab:tab3}. The detailed analyses are as follows:

\paragraph{Multimodal interactive encoder performs a primary backbone} When the multimodal interactive encoder is removed from \texttt{Amuse}, the final predictions lack interactive multimodal features. This ablation model shows performance drops across all question types (Audio Avg: -20.92\%, Visual Avg: -32.58\%, Audio-Visual Avg: -30.89\%, Overall Avg: -28.79\%), highlighting the essential role of the multimodal interactive encoder in all tasks: the interactive representation from early fusion forms the foundation of \texttt{Amuse}.

\paragraph{Rhythm encoder learns \textit{audio-related} musical characteristics for better counting} Excluding the temporal rhythm modeling in \texttt{Amuse} removes both the audio and visual rhythm encoders. This results in notable performance drops in audio-related questions (Audio Counting: -1.40\%, Audio Comparative: -2.28\%, Audio Average: -1.81\%) and a noticeable decline in Audio-Visual Counting questions (-1.43\%). These results indicate that the rhythm encoders are crucial for solving both audio-related and counting-related questions.

\paragraph{Music source encoder learns \textit{temporal} musical characteristics for final predictions} Excluding the temporal source modeling in \texttt{Amuse} removes both the audio and visual music source encoders, leading to noticeable performance drops in audio questions (Audio Counting: -5.94\%, Audio Average: -4.72\%) and significant decreases in Audio-Visual Temporal (-2.39\%) and Audio-Visual Counting questions (-5.55\%). These results indicate that the music source encoders effectively learn music source features and that aligning these sources with the temporal dimension helps answer temporal-related and counting-related questions.

\paragraph{Highway for musical RoIs helps visual tasks} Excluding the Music RoIs in \texttt{Amuse}, which removes the YoloV8 feature encoder in our implementations, results in noticeable performance declines in visual-related questions (Visual Avg: -7.39\%) and significant drops in Audio-Visual Existential (-6.00\%) and Audio-Visual Counting (-13.40\%). This indicates that the YoloV8 music feature encoder effectively captures musical regions of interest in video input, such as instruments, players, and conductors, thereby enhancing performance in visual-related question answering.

\subsection{Quantify Importance of Each Music Design to All Musical Tasks}
\label{subsec:exp-4}
\begin{figure}[htbp]
\begin{center}
\includegraphics[width=1.0\linewidth]{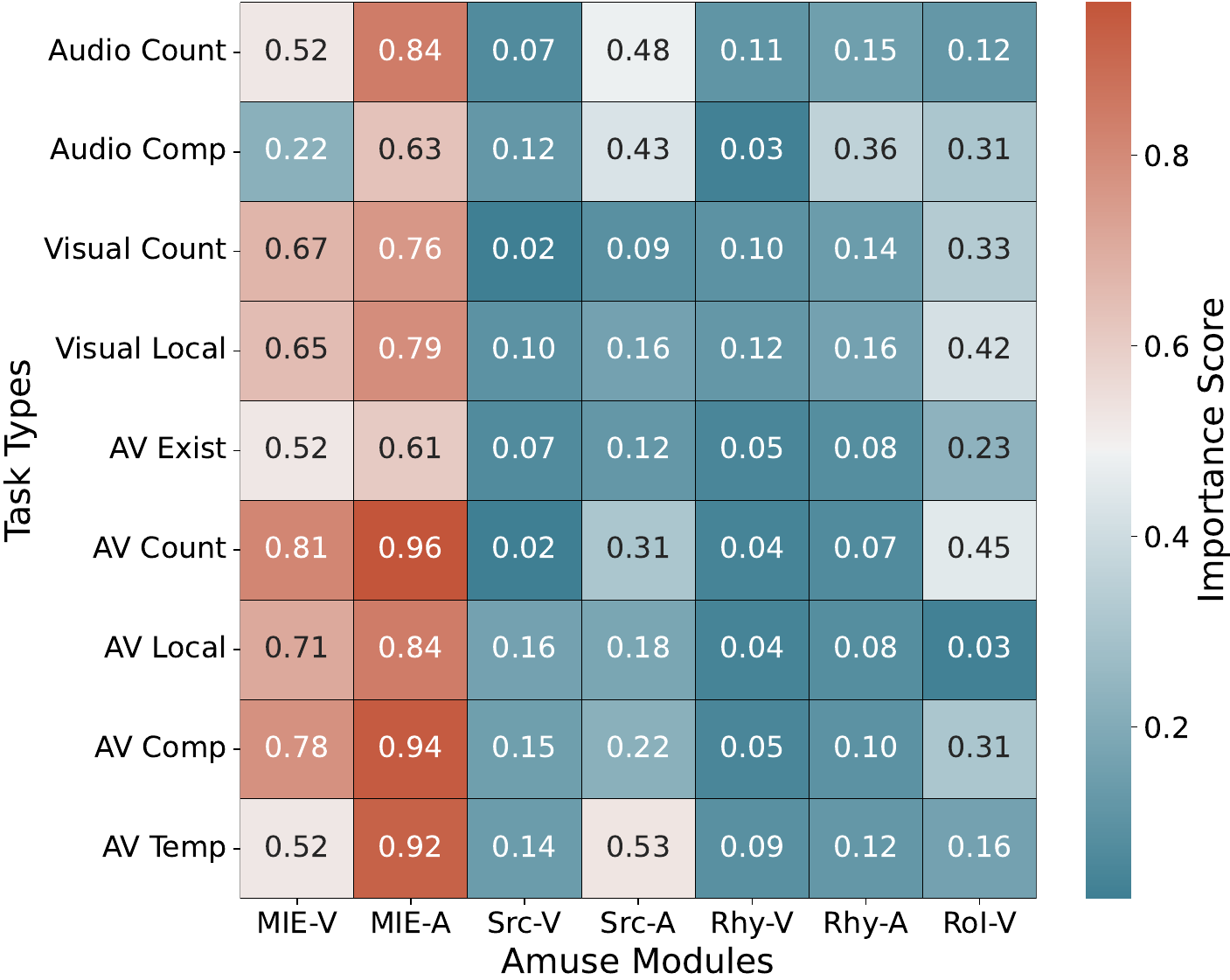}
\end{center}
\vspace{-0.1cm}
\caption{Distribution of importance scores for different modules (x-axis) in \texttt{Amuse} across various question types (y-axis). Each score ranges from 0 to 1, indicating the encoder significance as calculated by the attention layer for each question type in Music AVQA dataset. Higher scores reflect greater contributions to the final answers. Our results are averaged over three separate runs.}
\label{fig:vis}
\end{figure}
We quantify the importance score of each musical module in \texttt{Amuse} to different types of questions, as Fig.\ref{fig:vis} demonstrated. When iterating \texttt{Amuse} on each type of question, outputs from each sub-module (visual and audio transformers of the multimodal interactive encoder, temporal music source encoder, and temporal rhythm encoder, as well as the YoloV8 highway of musical RoIs) are combined with the question representation as inputs for the cross-modal attention layer. The calculated results from the cross-modal attention layers reflect the importance of each musical module's output concerning the semantics of the current question.

The visual transformers of the temporal source and rhythm encoders have a relatively minor impact on \texttt{Amuse}'s final inference outcomes. In contrast, the YoloV8 music element highway significantly contributes to the final predictions for visual questions, indicating that visual elements are critical for accurate visual question answering.
The temporal source encoder's audio transformer plays a substantial role in temporal and audio-related questions, indicating that audio features are crucial for handling questions that involve temporal dynamics and sound.
The multimodal interactive encoder, comprising visual and audio transformers, is a fundamental module for \texttt{Amuse}, exerting the most significant influence on its inference process. This suggests integrating visual and audio information is essential for the model's overall performance.

These observations enhance the transparency of \texttt{Amuse}'s prediction behavior concerning multimodal-related questions, providing insights into how different components contribute to the model's decision-making process. In conjunction with the results described in \S\ref{subsec:exp-5}, we obtain similar results in two experiments focusing on the modeling reasoning. Hence, \texttt{Amuse} demonstrates a foundational capability in modeling reasoning within AVQA, and its final feature mappings exhibit approximate linearity.

\section{Conclusion}
We introduce \texttt{Amuse}, the first framework specifically designed for Audio-Visual Question Answering in music. \texttt{Amuse} effectively addresses the main challenges in this field: (i) the lack of early-stage fusion for interconnected multimodal representation, (ii) the under-utilization of musical characteristics along the temporal dimension, and (iii) insufficient supervision for modeling rhythm and sources. By overcoming these challenges, \texttt{Amuse} provides valuable insights for future research on musical performance and multimodal understanding with continuous and dense audio.

\section{Limitations}

Although our approach succeeds on the Music AVQA dataset v1 and v2, it may not directly generalize to other types of tasks or datasets. The audio-specific information we leveraged, represented by rhythm and source in this task, may require rethinking for different contexts. For instance, if the audio does not involve musical instruments, the methods for extracting relevant audio-specific features would need to be adapted. The categories used for annotation depend on the specific dataset, and we may need to develop self-supervised algorithms to discover pertinent information in new datasets.

Moreover, the current framework's use of detailed and context-specific annotations highlights a potential consideration in scalability and adaptability to other non-music domains. Future work can explore more generalized approaches and the integration of self-supervised learning techniques to enhance the model's ability to adapt to diverse datasets and tasks without extensive manual annotations while considering sufficient multimodal representation interaction and audio-specific semantics.

\section*{Ethics Statement}
We have not identified any ethical concerns directly related to this study.

\section*{Acknowledgment}
This project is supported by the Department of Defense’s Congressionally Directed Medical Research Programs (DOD CDMRP) Award HT9425-23-1-0267.

\bibliography{ref}

\begin{thebibliography}{41}
\providecommand{\natexlab}[1]{#1}

\bibitem[{Agostinelli et~al.(2023)Agostinelli, Denk, Borsos, Engel, Verzetti, Caillon, Huang, Jansen, Roberts, Tagliasacchi et~al.}]{agostinelli2023musiclm}
Andrea Agostinelli, Timo~I Denk, Zal{\'a}n Borsos, Jesse Engel, Mauro Verzetti, Antoine Caillon, Qingqing Huang, Aren Jansen, Adam Roberts, Marco Tagliasacchi, et~al. 2023.
\newblock Musiclm: Generating music from text.
\newblock \emph{arXiv preprint arXiv:2301.11325}.

\bibitem[{Antol et~al.(2015)Antol, Agrawal, Lu, Mitchell, Batra, Zitnick, and Parikh}]{antol2015vqa}
Stanislaw Antol, Aishwarya Agrawal, Jiasen Lu, Margaret Mitchell, Dhruv Batra, C~Lawrence Zitnick, and Devi Parikh. 2015.
\newblock Vqa: Visual question answering.
\newblock In \emph{International Conference on Computer Vision}.

\bibitem[{Chen et~al.(2020)Chen, Xie, Vedaldi, and Zisserman}]{9053174}
Honglie Chen, Weidi Xie, Andrea Vedaldi, and Andrew Zisserman. 2020.
\newblock Vggsound: A large-scale audio-visual dataset.
\newblock In \emph{International Conference on Acoustics, Speech and Signal Processing}.

\bibitem[{Chen et~al.(2022)Chen, Du, Zhu, Ma, Berg-Kirkpatrick, and Dubnov}]{chen2022hts}
Ke~Chen, Xingjian Du, Bilei Zhu, Zejun Ma, Taylor Berg-Kirkpatrick, and Shlomo Dubnov. 2022.
\newblock Hts-at: A hierarchical token-semantic audio transformer for sound classification and detection.
\newblock In \emph{International Conference on Acoustics, Speech and Signal Processing}.

\bibitem[{Copet et~al.(2024)Copet, Kreuk, Gat, Remez, Kant, Synnaeve, Adi, and D{\'e}fossez}]{copet2024simple}
Jade Copet, Felix Kreuk, Itai Gat, Tal Remez, David Kant, Gabriel Synnaeve, Yossi Adi, and Alexandre D{\'e}fossez. 2024.
\newblock Simple and controllable music generation.
\newblock In \emph{Advances in Neural Information Processing Systems}.

\bibitem[{Dai et~al.(2022)Dai, Yu, and Dannenberg}]{dai2022missing}
Shuqi Dai, Huiran Yu, and Roger~B Dannenberg. 2022.
\newblock What is missing in deep music generation? a study of repetition and structure in popular music.
\newblock \emph{arXiv preprint arXiv:2209.00182}.

\bibitem[{Diao et~al.(2023)Diao, Cheng, and Cheng}]{diao2023av}
Xingjian Diao, Ming Cheng, and Shitong Cheng. 2023.
\newblock Av-maskenhancer: Enhancing video representations through audio-visual masked autoencoder.
\newblock In \emph{International Conference on Tools with Artificial Intelligence}.

\bibitem[{Dosovitskiy et~al.(2021)Dosovitskiy, Beyer, Kolesnikov, Weissenborn, Zhai, Unterthiner, Dehghani, Minderer, Heigold, Gelly et~al.}]{dosovitskiy2020image}
Alexey Dosovitskiy, Lucas Beyer, Alexander Kolesnikov, Dirk Weissenborn, Xiaohua Zhai, Thomas Unterthiner, Mostafa Dehghani, Matthias Minderer, Georg Heigold, Sylvain Gelly, et~al. 2021.
\newblock An image is worth 16x16 words: Transformers for image recognition at scale.
\newblock In \emph{International Conference on Learning Representations}.

\bibitem[{Duan et~al.(2024)Duan, Xia, Mingze, Tang, Zhu, and Zhao}]{duan2024cross}
Haoyi Duan, Yan Xia, Zhou Mingze, Li~Tang, Jieming Zhu, and Zhou Zhao. 2024.
\newblock Cross-modal prompts: Adapting large pre-trained models for audio-visual downstream tasks.
\newblock In \emph{Advances in Neural Information Processing Systems}.

\bibitem[{Fayek and Johnson(2020)}]{fayek2020temporal}
Haytham~M. Fayek and Justin Johnson. 2020.
\newblock Temporal reasoning via audio question answering.
\newblock \emph{Transactions on Audio, Speech, and Language Processing}.

\bibitem[{Garcia et~al.(2020)Garcia, Otani, Chu, and Nakashima}]{garcia2020knowit}
Noa Garcia, Mayu Otani, Chenhui Chu, and Yuta Nakashima. 2020.
\newblock Knowit vqa: Answering knowledge-based questions about videos.
\newblock In \emph{AAAI Conference on Artificial Intelligence}.

\bibitem[{Huang et~al.(2024)Huang, Li, Yang, Shi, Chang, Ye, Wu, Hong, Huang, Liu et~al.}]{huang2024audiogpt}
Rongjie Huang, Mingze Li, Dongchao Yang, Jiatong Shi, Xuankai Chang, Zhenhui Ye, Yuning Wu, Zhiqing Hong, Jiawei Huang, Jinglin Liu, et~al. 2024.
\newblock Audiogpt: Understanding and generating speech, music, sound, and talking head.
\newblock In \emph{AAAI Conference on Artificial Intelligence}.

\bibitem[{Kingma and Ba(2015)}]{kingma2015adam}
Diederick~P Kingma and Jimmy Ba. 2015.
\newblock Adam: A method for stochastic optimization.
\newblock In \emph{International Conference on Learning Representations}.

\bibitem[{Kong et~al.(2023)Kong, Chen, Liu, Du, Berg-Kirkpatrick, Dubnov, and Plumbley}]{kong2023universal}
Qiuqiang Kong, Ke~Chen, Haohe Liu, Xingjian Du, Taylor Berg-Kirkpatrick, Shlomo Dubnov, and Mark~D. Plumbley. 2023.
\newblock Universal source separation with weakly labelled data.
\newblock \emph{arXiv preprint arXiv:2305.07447}.

\bibitem[{Lei et~al.(2018)Lei, Yu, Bansal, and Berg}]{lei2018tvqa}
Jie Lei, Licheng Yu, Mohit Bansal, and Tamara~L Berg. 2018.
\newblock Tvqa: Localized, compositional video question answering.
\newblock In \emph{Conference on Empirical Methods in Natural Language Processing}.

\bibitem[{Li et~al.(2023{\natexlab{a}})Li, Hou, and Hu}]{li2023progressive}
Guangyao Li, Wenxuan Hou, and Di~Hu. 2023{\natexlab{a}}.
\newblock Progressive spatio-temporal perception for audio-visual question answering.
\newblock In \emph{International Conference on Multimedia}.

\bibitem[{Li et~al.(2022)Li, Wei, Tian, Xu, Wen, and Hu}]{li2022learning}
Guangyao Li, Yake Wei, Yapeng Tian, Chenliang Xu, Ji-Rong Wen, and Di~Hu. 2022.
\newblock Learning to answer questions in dynamic audio-visual scenarios.
\newblock In \emph{Conference on Computer Vision and Pattern Recognition}.

\bibitem[{Li et~al.(2023{\natexlab{b}})Li, Xu, and Hu}]{li2023multi}
Guangyao Li, Yixin Xu, and Di~Hu. 2023{\natexlab{b}}.
\newblock Multi-scale attention for audio question answering.
\newblock In \emph{Annual Conference of the International Speech Communication Association}.

\bibitem[{Lin et~al.(2023)Lin, Sung, Lei, Bansal, and Bertasius}]{lin2023vision}
Yan-Bo Lin, Yi-Lin Sung, Jie Lei, Mohit Bansal, and Gedas Bertasius. 2023.
\newblock Vision transformers are parameter-efficient audio-visual learners.
\newblock In \emph{Conference on Computer Vision and Pattern Recognition}.

\bibitem[{Lipping et~al.(2022)Lipping, Sudarsanam, Drossos, and Virtanen}]{lipping2022clotho}
Samuel Lipping, Parthasaarathy Sudarsanam, Konstantinos Drossos, and Tuomas Virtanen. 2022.
\newblock Clotho-aqa: A crowdsourced dataset for audio question answering.
\newblock In \emph{European Signal Processing Conference}.

\bibitem[{Liu et~al.(2023)Liu, Xie, Gao, and Yu}]{liu2023parameter}
Hongye Liu, Xianhai Xie, Yang Gao, and Zhou Yu. 2023.
\newblock Parameter-efficient transfer learning for audio-visual-language tasks.
\newblock In \emph{International Conference on Multimedia}.

\bibitem[{Liu et~al.(2024)Liu, Dong, and Zhang}]{liu2024tackling}
Xiulong Liu, Zhikang Dong, and Peng Zhang. 2024.
\newblock Tackling data bias in music-avqa: Crafting a balanced dataset for unbiased question-answering.
\newblock In \emph{Winter Conference on Applications of Computer Vision}.

\bibitem[{Liu et~al.(2022)Liu, Hu, Lin, Yao, Xie, Wei, Ning, Cao, Zhang, Dong et~al.}]{liu2022swin}
Ze~Liu, Han Hu, Yutong Lin, Zhuliang Yao, Zhenda Xie, Yixuan Wei, Jia Ning, Yue Cao, Zheng Zhang, Li~Dong, et~al. 2022.
\newblock Swin transformer v2: Scaling up capacity and resolution.
\newblock In \emph{Conference on Computer Vision and Pattern Recognition}.

\bibitem[{Lu et~al.(2023)Lu, Xu, Kang, Yu, Xing, Tan, and Bian}]{lu2023musecoco}
Peiling Lu, Xin Xu, Chenfei Kang, Botao Yu, Chengyi Xing, Xu~Tan, and Jiang Bian. 2023.
\newblock Musecoco: Generating symbolic music from text.
\newblock \emph{arXiv preprint arXiv:2306.00110}.

\bibitem[{Ngiam et~al.(2011)Ngiam, Khosla, Kim, Nam, Lee, and Ng}]{ngiam2011multimodal}
Jiquan Ngiam, Aditya Khosla, Mingyu Kim, Juhan Nam, Honglak Lee, and Andrew~Y Ng. 2011.
\newblock Multimodal deep learning.
\newblock In \emph{International Conference on Machine Learning}.

\bibitem[{Ravi et~al.(2023)Ravi, Chinchure, Sigal, Liao, and Shwartz}]{ravi2023vlc}
Sahithya Ravi, Aditya Chinchure, Leonid Sigal, Renjie Liao, and Vered Shwartz. 2023.
\newblock Vlc-bert: visual question answering with contextualized commonsense knowledge.
\newblock In \emph{Winter Conference on Applications of Computer Vision}.

\bibitem[{Saito et~al.(2024)Saito, Sohn, Lee, and Ushiku}]{saito2024unsupervised}
Kuniaki Saito, Kihyuk Sohn, Chen-Yu Lee, and Yoshitaka Ushiku. 2024.
\newblock Unsupervised llm adaptation for question answering.
\newblock \emph{arXiv preprint arXiv:2402.12170}.

\bibitem[{Shen et~al.(2024)Shen, Song, Tan, Li, Lu, and Zhuang}]{shen2024hugginggpt}
Yongliang Shen, Kaitao Song, Xu~Tan, Dongsheng Li, Weiming Lu, and Yueting Zhuang. 2024.
\newblock Hugginggpt: Solving ai tasks with chatgpt and its friends in hugging face.
\newblock \emph{Advances in Neural Information Processing Systems}.

\bibitem[{Srivastava and Salakhutdinov(2012)}]{srivastava2012multimodal}
Nitish Srivastava and Russ~R. Salakhutdinov. 2012.
\newblock Multimodal learning with deep boltzmann machines.
\newblock In \emph{Advances in Neural Information Processing Systems}.

\bibitem[{Sudarsanam and Virtanen(2023)}]{sudarsanam2023attention}
Parthasaarathy Sudarsanam and Tuomas Virtanen. 2023.
\newblock Attention-based methods for audio question answering.
\newblock In \emph{European Signal Processing Conference}.

\bibitem[{Tian et~al.(2020)Tian, Li, and Xu}]{tian2020unified}
Yapeng Tian, Dingzeyu Li, and Chenliang Xu. 2020.
\newblock Unified multisensory perception: Weakly-supervised audio-visual video parsing.
\newblock In \emph{European Conference on Computer Vision}.

\bibitem[{Ultralytics(2023)}]{Ultralytics2023yolov8}
Ultralytics. 2023.
\newblock Yolov8.
\newblock \url{https://github.com/ultralytics/ultralytics}.
\newblock Accessed: 2023-05-26.

\bibitem[{Vaswani et~al.(2017)Vaswani, Shazeer, Parmar, Uszkoreit, Jones, Gomez, Kaiser, and Polosukhin}]{NIPS2017_3f5ee243}
Ashish Vaswani, Noam Shazeer, Niki Parmar, Jakob Uszkoreit, Llion Jones, Aidan~N Gomez, \L~ukasz Kaiser, and Illia Polosukhin. 2017.
\newblock Attention is all you need.
\newblock In \emph{Advances in Neural Information Processing Systems}.

\bibitem[{Wang et~al.(2024)Wang, Li, Huang, and Rahmani}]{wang2024healthq}
Ziyu Wang, Hao Li, Di~Huang, and Amir~M. Rahmani. 2024.
\newblock Healthq: Unveiling questioning capabilities of llm chains in healthcare conversations.
\newblock \emph{arXiv preprint arXiv:2409.19487}.

\bibitem[{Yang et~al.(2022)Yang, Wang, Duan, Chen, Hou, Jin, and Zhu}]{yang2022avqa}
Pinci Yang, Xin Wang, Xuguang Duan, Hong Chen, Runze Hou, Cong Jin, and Wenwu Zhu. 2022.
\newblock Avqa: A dataset for audio-visual question answering on videos.
\newblock In \emph{International Conference on Multimedia}.

\bibitem[{Yu et~al.(2024)Yu, Cho, Yadav, and Bansal}]{yu2024self}
Shoubin Yu, Jaemin Cho, Prateek Yadav, and Mohit Bansal. 2024.
\newblock Self-chained image-language model for video localization and question answering.
\newblock In \emph{Advances in Neural Information Processing Systems}.

\bibitem[{Yu et~al.(2019)Yu, Xu, Yu, Yu, Zhao, Zhuang, and Tao}]{yu2019activitynet}
Zhou Yu, Dejing Xu, Jun Yu, Ting Yu, Zhou Zhao, Yueting Zhuang, and Dacheng Tao. 2019.
\newblock Activitynet-qa: A dataset for understanding complex web videos via question answering.
\newblock In \emph{AAAI Conference on Artificial Intelligence}.

\bibitem[{Yun et~al.(2021)Yun, Yu, Yang, Lee, and Kim}]{yun2021pano}
Heeseung Yun, Youngjae Yu, Wonsuk Yang, Kangil Lee, and Gunhee Kim. 2021.
\newblock Pano-avqa: Grounded audio-visual question answering on 360deg videos.
\newblock In \emph{International Conference on Computer Vision}.

\bibitem[{Zhao et~al.(2019)Zhao, Gan, Ma, and Torralba}]{zhao2019sound}
Hang Zhao, Chuang Gan, Wei-Chiu Ma, and Antonio Torralba. 2019.
\newblock The sound of motions.
\newblock In \emph{International Conference on Computer Vision}.

\bibitem[{Zhao et~al.(2018)Zhao, Gan, Rouditchenko, Vondrick, McDermott, and Torralba}]{zhao2018sound}
Hang Zhao, Chuang Gan, Andrew Rouditchenko, Carl Vondrick, Josh McDermott, and Antonio Torralba. 2018.
\newblock The sound of pixels.
\newblock In \emph{European Conference on Computer Vision}.

\bibitem[{Zhuang et~al.(2023)Zhuang, Yu, Wang, Sun, and Zhang}]{zhuang2023toolqa}
Yuchen Zhuang, Yue Yu, Kuan Wang, Haotian Sun, and Chao Zhang. 2023.
\newblock Toolqa: A dataset for llm question answering with external tools.
\newblock \emph{Advances in Neural Information Processing Systems}.

\end{thebibliography}
\end{document}